\documentclass[preprint,12pt]{elsarticle}

\usepackage{amssymb}

\usepackage{algorithmic}
\usepackage{algorithm}
\usepackage{array}

\usepackage{amsmath}
\usepackage{url}
\usepackage{graphicx}
\usepackage{booktabs}
\usepackage{multirow}
\usepackage{caption}
\usepackage{subcaption}
\usepackage{hyperref}
\usepackage{amsthm}
\usepackage{natbib}
\usepackage{xcolor}

\newtheorem{lemma}{Lemma}

\newtheorem{assumption}{Assumption}

\journal{Nuclear Physics B}

\begin{document}

\begin{frontmatter}

\title{Contrastive Adapter Training and Consensus Knowledge Distillation for Multi-Source-Free Domain Adaptation in Skin Cancer Diagnosis}

\author[label1]{Ali Abedi}
\author[label1]{Q. M. Jonathan Wu}
\author[label1]{Ning Zhang}

\affiliation[label1]{organization={University of Windsor, Department of Electrical and Computer Engineering}}

\author[label2]{Farhad Pourpanah}

\affiliation[label2]{organization={Queen's University, Department of Electrical and Computer Engineering}}

\begin{abstract}
{\color{black}
Skin cancer diagnosis, particularly the differentiation of melanoma from benign nevi, is a vital yet challenging task due to the visual similarity between lesions. Although deep learning models such as convolutional neural networks (CNNs) and vision transformers (ViTs) have demonstrated promising performance, their effectiveness often deteriorates when applied to data from heterogeneous clinical sources. While conventional domain adaptation methods address domain shift, they require access to source data during adaptation, which is often infeasible due to privacy regulations. Multi-source-free unsupervised domain adaptation (MSFDA) addresses this limitation by leveraging multiple labeled source domains to generalize to an unlabeled target domain without requiring access to source data, making it suitable for privacy-sensitive medical settings. However, existing MSFDA methods rely on full backbone fine-tuning, leading to catastrophic forgetting and overfitting on small clinical datasets, and address domain shift at the aggregation stage without establishing a shared domain-invariant feature space. Furthermore, their reliance on hard pseudo-labels or confidence-weighted aggregation introduces noisy supervision signals under domain shift. To address these limitations, we propose CAT-CKD, consisting of two components: (1) contrastive adapter training (CAT), which trains lightweight ConvPass adapters within a frozen ViT backbone using supervised contrastive learning (SCL) to establish a shared domain-invariant feature space before source-specific model training, and (2) consensus knowledge distillation (CKD), which aggregates logits from multiple source models into a consensus supervisory signal and adapts a student model on unlabeled target data using KL divergence. Experiments on five publicly available skin lesion datasets show that CAT-CKD achieves an average AUROC of 86.1\%, outperforming existing MSFDA methods while requiring only 4.3M trainable parameters. The code for this paper is available at \url{https://github.com/A-Abedi/CAT_CKD}.}
\end{abstract}


\end{frontmatter}

\section{Introduction}
\label{sec:introduction}
Melanoma is one of the most aggressive and potentially fatal forms of skin cancer, responsible for a large number of skin cancer-related deaths worldwide \cite{tiwari_survey_2024, kreouzi_deep_2024}. Early and accurate diagnosis of melanoma is crucial, as it is highly treatable in its initial stages but can rapidly metastasize if left undetected \cite{melarkode_ai-powered_2023}. Differentiating melanoma from benign skin lesions such as nevi is a fundamental challenge in dermatology due to their similar visual appearance \cite{fink_diagnostic_2020, jackson_differentiating_2025}. This motivates the development of computer-aided diagnostic systems to assist clinicians in improving diagnostic accuracy.

In recent years, deep learning techniques, such as convolutional neural networks (CNNs) \cite{he_deep_2016, howard_mobilenets_2017, huang_densely_2018} and vision transformers (ViTs) \cite{dosovitskiy_image_2021}, have shown promising results in medical image analysis, including skin cancer diagnosis \cite{fink_diagnostic_2020, kreouzi_deep_2024}. {\color{black}However, the limited availability of annotated data drives researchers to rely on existing labeled datasets collected from multiple clinical centers, as annotation requires scarce domain expertise and is prohibitively time-consuming \cite{zhang_collaborative_2020}.} Nevertheless, a major obstacle remains: the domain shift problem. Variations in imaging devices, acquisition protocols, and patient demographics across centers introduce discrepancies between source and target domains, which substantially degrade the generalizability of deep learning models when deployed in previously unseen clinical data \cite{wang_deep_2018, redko_advances_2019, han_learning_2022, cai_multi-source_2025}. {\color{black}Compounding this challenge, privacy regulations and institutional policies \cite{jiang_privacy-preserving_2026} typically prevent source data from being revisited once a model is deployed to a new clinical environment \cite{zhang_cluster-fair-fl_2026}.}

{\color{black}
Multi-source-free unsupervised domain adaptation (MSFDA) addresses domain shift by leveraging labeled data from multiple source domains to generalize to an unlabeled target domain without requiring access to source data during adaptation. This setup aligns better with real-world clinical settings, where privacy regulations often prevent access to source datasets during deployment \cite{dong_confident_2021, cai_make_2026}. While promising, existing MSFDA methods suffer from two key limitations. First, they rely on full backbone fine-tuning during adaptation, which causes catastrophic forgetting of pretrained representations and overfitting on the small, imbalanced datasets typical of clinical settings. Second, they attempt to reconcile domain shift at the aggregation or adaptation stage \cite{zhang_more_2025}, without first establishing a shared domain-invariant backbone, which if available, could be frozen during source-specific model training, reducing both overfitting and computational overhead. Furthermore, their reliance on hard pseudo-labels that directly propagate teacher errors \cite{ahmed_unsupervised_2021}, or confidence-weighted aggregation that amplifies errors from overconfident teachers \cite{li_agile_2024, pei_evidential_2024}, introduces noisy supervision signals that are unreliable under the domain shift conditions inherent to multi-center clinical data.
}

{\color{black}To address the challenge of noisy supervision under domain shift,} knowledge distillation (KD) \cite{hinton_distilling_2015} can facilitate the transfer of information from well-trained source models to a target model, which enables effective learning without direct access to the source data \cite{nguyen-meidine_joint_2020, nguyen-meidine_knowledge_2021}. Using multiple teacher models allows KD to mitigate individual source biases and provide more robust supervision signals for the student model. The success of this process depends heavily on the quality of the teacher models, which must be carefully fine-tuned to ensure they provide accurate and consistent guidance.

{\color{black}
Addressing the absence of a shared domain-invariant backbone requires a training stage that leverages all available labeled source data jointly with an objective that explicitly encourages same-class samples from different domains to be close together in the feature space. However, directly fine-tuning large pretrained ViTs for this purpose is computationally expensive and prone to overfitting and catastrophic forgetting, particularly when labeled data are scarce and imbalanced across clinical centers. Parameter-efficient fine-tuning (PEFT) methods \cite{han_parameter-efficient_2024}, including adapters \cite{yang_aim_2023, jie_convolutional_2024} and low-rank adaptation (LoRA) \cite{hu_lora_2021, yuan_fulllora-at_2024}, offer a practical solution by updating only a small subset of parameters while keeping the backbone frozen, thereby preserving pretrained knowledge and reducing the risk of overfitting. This makes PEFT particularly well-suited for building a shared domain-invariant backbone in the MSFDA setting, where labeled data are limited and privacy constraints prevent repeated access to source data during deployment.}

{\color{black}
This paper introduces CAT-CKD, an MSFDA framework for skin cancer diagnosis that directly addresses the limitations of prior methods through two complementary components. First, we introduce \textit{\textbf{C}ontrastive \textbf{A}dapter \textbf{T}raining (CAT)}, a PEFT strategy that inserts convolutional bypass adapters (ConvPass) \cite{jie_convolutional_2024} into a frozen ViT backbone and trains them using supervised contrastive learning (SCL) \cite{khosla_supervised_2021}, which encourages intra-class compactness and inter-class separation across domains regardless of their origin, to establish a shared domain-invariant feature space before source-specific models are built. Second, we design a \textit{\textbf{C}onsensus \textbf{K}nowledge \textbf{D}istillation (CKD)} strategy, in which the logits from multiple trained source models are aggregated into a consensus supervision signal, which we show reduces logit-level variance compared to any single teacher, and used to adapt a student model on the unlabeled target domain via Kullback--Leibler (KL) divergence, avoiding the unreliable confidence estimates that plague existing aggregation strategies under domain shift. Because adaptation relies solely on trained source models, the proposed framework avoids direct access to source-domain images during deployment. Extensive experiments demonstrate that CAT-CKD provides a practical and parameter-efficient solution for source-free adaptation in skin cancer diagnosis under data-sharing constraints.

The main contributions of this work include:
\begin{itemize}
\item We propose CAT-CKD, an MSFDA framework for skin cancer diagnosis that explicitly establishes a shared domain-invariant backbone before source-specific models are built, directly addressing the representation-level and supervision-level limitations of existing methods.
\item We introduce CAT, which trains ConvPass adapters using supervised contrastive learning across source domains in the MSFDA setting, producing a shared domain-invariant feature space while avoiding catastrophic forgetting and overfitting.
\item We design CKD, which aggregates logits from multiple source models into a theoretically grounded consensus supervisory signal and adapts a student model on unlabeled target data using a KL-divergence objective, providing more stable supervision than existing aggregation strategies.
\item We conduct extensive experiments, ablation studies, and efficiency analyses on five public skin lesion datasets, demonstrating competitive performance with substantially fewer trainable parameters than existing methods.
\end{itemize}
}

The remainder of this paper is organized as follows. Section~\ref{sec:relatedworks} reviews related work in skin cancer diagnosis, domain adaptation, parameter-efficient fine tuning, and knowledge distillation. Section~\ref{sec:methodology} presents the problem formulation and details the proposed CAT-CKD framework. Section~\ref{sec:experiments} describes the experimental setup and discusses the results. Finally, Section~\ref{sec:conclusion} concludes the paper and outlines directions for future work.

\section{Related Work}
\label{sec:relatedworks}
\subsection{Skin Cancer Diagnosis}
Recent research has increasingly leveraged deep learning methods to distinguish melanoma from benign nevi. CNNs are the most widely used techniques, typically fine-tuned from {\color{black}ImageNet-pretrained architectures such as DenseNet \cite{huang_densely_2018}, which uses dense connections between layers to encourage feature reuse, ResNet \cite{he_deep_2016}, which introduces residual connections to enable training of very deep networks, and MobileNet \cite{howard_mobilenets_2017}, which employs depthwise separable convolutions for lightweight and efficient inference}. In \cite{fink_diagnostic_2020}, a pre-trained CNN model is fine-tuned on over 120,000 dermoscopic images to classify lesions as either melanoma or benign, using a softmax-generated malignancy probability threshold of 0.5. The study in \cite{kreouzi_deep_2024} fine-tunes four pretrained CNN architectures, including DenseNet121, ResNet50V2, NASNetMobile, and MobileNetV2, by applying extensive data augmentation to a dataset of 8,825 images. However, CNNs primarily focus on local features due to their limited receptive fields, which restricts their ability to capture long-range dependencies and global context \cite{mauricio_comparing_2023}.

To overcome the limitations of CNNs, recent studies have explored ViTs, which employ self-attention mechanisms \cite{vaswani_attention_2017} to model relationships between image regions. This design enables ViTs to integrate information across both local and long-range spatial dependencies. Their success across a wide range of vision tasks has generated growing interest in applying ViTs to skin cancer diagnosis. RETFound \cite{jackson_differentiating_2025} fine-tunes a ViT model to classify ultrawidefield color and autofluorescence fundoscopy images without the need for augmentation or handcrafted features. In \cite{aksoy_enhancing_2024}, architectures such as ConvNeXt, ViT-Base-16, and Swin V2 leverage self-attention and patch-based designs to enhance diagnostic performance. Additionally, GAN-MBViT \cite{wang_novel_2024} employs a dual-branch training strategy to boost feature learning by combining StyleGAN2-ADA for generating melanoma images with a ViT enhanced by BatchFormer for classification. Finally, ScATNet \cite{wu_scale-aware_2021} applies a multi-scale ViT to histopathology slides by using self-attention across image patches at different magnifications, enabling the effective fusion of features from various tissue scales. {\color{black}Despite these advances, existing skin cancer diagnosis methods typically do not address the domain shift that arises when models are deployed across heterogeneous clinical sources.}

\subsection{Unsupervised Domain Adaptation}
UDA aims to transfer knowledge from a labeled source domain to an unlabeled target domain with a different data distribution. DANN \cite{csurka_domain-adversarial_2017} learns domain-invariant and task-discriminative features by introducing a domain classifier connected through a gradient reversal layer. It reverses gradients during backpropagation to encourage indistinguishability between source and target domains. The method jointly trains the label predictor and domain classifier in an adversarial manner. ADDA \cite{tzeng_adversarial_2017} first learns a discriminative feature mapping from source data and then trains a separate target encoder to map target data into the same space through adversarial training against a domain discriminator. It uses unshared weights and a GAN-based loss to align source and target representations without requiring image generation or weight sharing. Beyond single-source UDA, many tasks require adaptation from multiple sources. {\color{black}MDAN \cite{zhao_adversarial_2018} derives new generalization bounds for multiple source domain adaptation that motivate jointly learning domain-invariant, task-discriminative features. Two variants are proposed: a hard version optimizing a worst-case bound and a soft version using log-sum-exp weighting to adaptively combine gradients from all source domains.}

While multi-source unsupervised domain adaptation requires access to source data, multi-source-free variants rely solely on trained source models and are therefore better suited to real-world scenarios.  
DECISION \cite{ahmed_unsupervised_2021} addresses MSFDA by leveraging only pre-trained source models and unlabeled target data to construct a target model through adaptive ensembling. It jointly optimizes model-specific feature extractors and combines weights through information maximization and pseudo-labeling. This design ensures that the target model performs at least as well as the best individual source model.
{\color{black}EAAF \cite{pei_evidential_2024} introduces two uncertainty measures: Evidential Prediction Uncertainty (EPU), which estimates how well each source model fits a given target sample to guide instance-level aggregation, and Evidential Adjacency-Consistent Uncertainty (EAU), which measures semantic consistency within local neighborhoods to enable reliable semantic propagation, all within a dual-branch Multinomial-Dirichlet network.}
Bi-ATEN \cite{li_agile_2024} learns instance-specific and domain-consistent ensemble weights through a bi-level attention mechanism. It requires only minimal tuning of the source bottlenecks. {\color{black} While these methods demonstrate promising adaptation performance, they either require source data during adaptation or address domain shift only at the aggregation stage.}

\subsection{Parameter-Efficient Fine Tuning}
PEFT methods aim to adapt large pretrained models to downstream tasks using a small number of trainable parameters. Early PEFT strategies for language models have inspired several adaptation techniques for vision models. {\color{black}Adapters \cite{houlsby_parameter-efficient_2019} insert lightweight bottleneck modules between transformer layers, LoRA \cite{hu_lora_2021} decomposes weight updates into low-rank matrices to reduce trainable parameters, and Prefix-Tuning \cite{li_prefix-tuning_2021} prepends trainable tokens to the input sequence to guide model behavior, all while keeping the backbone frozen. Vision-specific PEFT variants have also shown competitive or superior performance to full fine-tuning, especially in low-data regimes. VPT \cite{jia_visual_2022} prepends learnable prompt tokens to the input of each transformer layer to adapt the model to downstream tasks. AdaptFormer \cite{chen_adaptformer_2022} replaces the MLP layers in ViT with lightweight adapter modules for scalable visual recognition. NOAH \cite{zhang_neural_2024} searches for optimal combinations of PEFT modules through neural architecture search to maximize adaptation performance.} While methods like Med-SA \cite{wu_medical_2025} apply PEFT principles to interactive segmentation in medical imaging, they rely on user prompts and are not designed for multi-source-free domain adaptation. Most PEFT approaches also remain language-oriented, inheriting architectural biases suitable for NLP, such as sequence handling and weak spatial priors. Convpass \cite{jie_convolutional_2024} addresses this issue by proposing a vision-oriented PEFT method. It introduces convolutional bypasses in parallel with the multi-head self-attention (MHSA) and MLP blocks within ViT layers. These modules restore spatial locality by applying convolutional transformations to the flattened patch embeddings to bridge the gap between CNNs and transformers.

\subsection{Knowledge Distillation}
KD has become a widely adopted approach for compressing deep models by transferring knowledge from a large, well-performing teacher model to a smaller student model. Hinton \textit{et al.} \cite{hinton_distilling_2015} introduced KD as a technique to transfer the generalization ability of a large or ensemble model (the teacher) to a smaller model (the student) by training the student on soft targets. KDCL \cite{guo_online_2020} is an online KD framework where multiple student networks collaboratively learn by sharing dynamically generated ensemble soft targets. This approach enhances generalization and does not rely on a pre-trained teacher. Attention transfer \cite{zagoruyko_paying_2017} is a KD method that improves student CNNs by forcing them to mimic the teacher’s spatial attention maps. While most early works used a single teacher, recent methods leverage multiple teachers to provide a broader supervision signal to the student. For example, AE-KD \cite{du_agree_2020} formulates ensemble KD as a multi-objective optimization in gradient space. It adaptively weights teacher contributions based on their agreement to guide the student more effectively. MMKD \cite{zhang_adaptive_2023} introduces a meta-weight network that adaptively combines multiple teachers' logits and feature representations at the instance level using second-order optimization to ensure that the student receives diverse yet compatible knowledge tailored to its learning capacity. FEED \cite{park_feed_2019} distills knowledge from multiple teachers at the feature-map level by introducing separate non-linear transformation layers for each teacher. This design allows the student network to align and absorb ensemble feature representations either in parallel or through sequential training.

\section{Problem Formulation and Proposed Solution}
\label{sec:methodology}
\subsection{Problem Formulation}
Consider $M$ source models, each possessing a dataset derived from a unique data distribution and distinct source domain $\mathcal{D}_{i}^{S} = \{x_{ij}, y_{ij}\}_{j=1}^{N_i}$, where $x_{ij} \in \mathcal{X}_i$ and $y_{ij} \in \mathcal{Y}_i$ represent the $j^{th}$ input and its corresponding label for the $i^{th}$ source domain. These datasets are distributionally diverse, with each dataset sampled from a distinct joint distribution \( P_i(X, Y) \), such that \( P_i \neq P_k \) for \( i \neq k \). $N_i$ denotes the number of samples of the $i^{th}$ source domain ($i = 1, 2, \ldots, M$). 

Our objective is to train a model that performs well on an unlabeled target domain \( \mathcal{D}^T = \{x_j^T\}_{j=1}^{N_T} \), where \( N_T \) denotes the number of target samples, by leveraging a set of source models trained on labeled source domains. Each source model \( f_i \) is trained on its respective labeled dataset by minimizing the following objective:
\begin{equation}
\min \mathcal{L}_i(f_i) = \frac{1}{N_i} \sum_{j=1}^{N_i} \ell(f_i(x_{ij}), y_{ij}),
\end{equation}
where \(\ell(\cdot)\) denotes the binary cross-entropy loss. Once trained, these source models are used in the adaptation phase to guide the learning of a domain-adapted model without requiring access to the original source data.

\begin{figure*}[!t]
\centerline{\includegraphics[width=\textwidth]{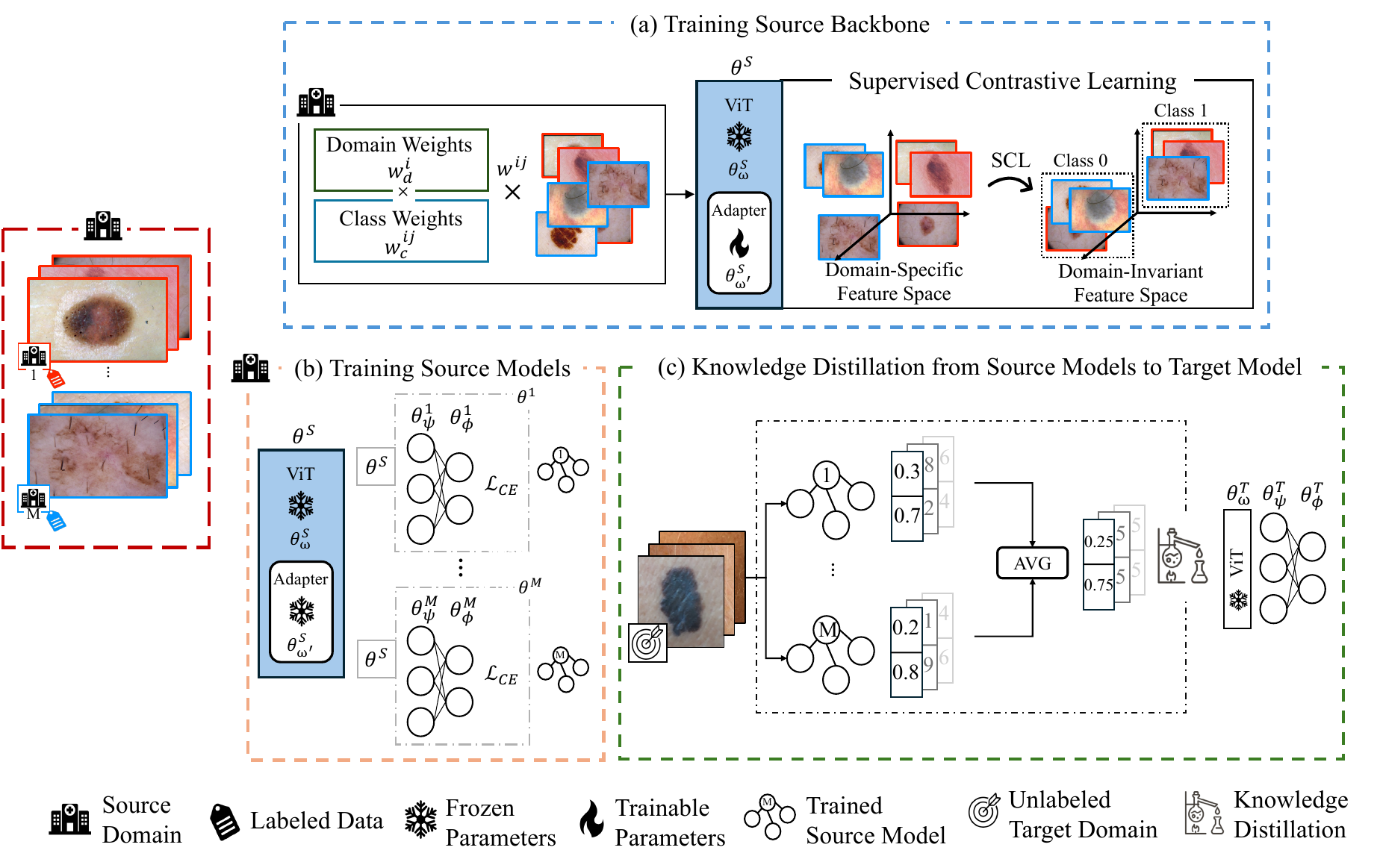}}
\caption{{\color{black}Overview of the proposed CAT-CKD framework. (a) CAT: ConvPass adapters are trained on pooled source-domain data using SCL, while the ViT backbone remains frozen to form a shared domain-invariant feature space. Domain and class imbalance are addressed via weighted sampling. (b) Source model training: Each domain independently trains a bottleneck and classification head on the frozen backbone with binary CE loss. (c) CKD: Source model logits on unlabeled target samples are averaged to form a consensus signal, which supervises the student model through KL divergence. Samples with the same border color belong to the same source domain.\protect\footnotemark}}
\label{fig:ModelOverview}
\end{figure*}
\footnotetext{Distillation icon created by Sir.Vector - Flaticon.}

\subsection{Model Overview}
Figure~\ref{fig:ModelOverview} provides an overview of the proposed CAT-CKD framework for skin-cancer diagnosis. The framework operates on data from multiple source domains that may exhibit domain shift due to variations in imaging protocols, patient demographics, and annotation practices.

CAT-CKD consists of two stages. To obtain class-discriminative and domain-invariant representations, CAT (Section~\ref{sec:CAT}) trains a ViT-based feature extractor $\theta^{S}$ using labeled data from all source domains, optimizing only the adapter parameters $\theta_{\omega'}^{S}$ with SCL \cite{khosla_supervised_2021} while keeping the ViT \cite{dosovitskiy_image_2021} backbone parameters $\theta_{\omega}^{S}$ frozen. Domain- and class-level weighting is applied to reduce the impact of imbalance. The resulting backbone is then frozen and shared across the $M$ source models; each source model attaches a domain-specific parameter set $\theta^{i}=\{\theta_{\psi}^{i},\theta_{\phi}^{i}\}$, where $\theta_{\psi}^{i}$ and $\theta_{\phi}^{i}$ denote the bottleneck and classification head, respectively, and is trained separately using only its local labeled data.

In the second stage, the trained source models act as an ensemble of teachers for an unlabeled target domain. CKD (Section~\ref{sec:CKD}) aggregates their predictions on target samples by averaging to form a consensus supervisory signal. A student model is then trained by aligning its outputs with this consensus via KL divergence \cite{hinton_distilling_2015} by distilling knowledge from all sources into a single target-adapted model. Since adaptation relies only on trained source models, the framework avoids direct access to source-domain data during deployment, aligning with data-sharing constraints in medical applications.

\subsection{Contrastive Adapter Training}
\label{sec:CAT}
 CAT is a PEFT approach that incorporates adapters into the ViT architecture to mitigate the catastrophic forgetting challenge.  These adapters are convolutional bypass modules \cite{jie_convolutional_2024} added in parallel to the existing MHSA and MLP blocks which allows efficient adaptation while preserving the original backbone structure. The modified transformation process in ViT with adapters is formulated as:
\begin{multline}
    X' \leftarrow X + \text{MHSA/MLP}(\text{LN}(X)) + \text{Adapter}(\text{LN}(X)), 
\end{multline}
where $\text{LN}(\cdot)$ is the layer normalization function, MHSA/MLP($\cdot$) represents the self-attention and feed-forward layers in ViT, and $\text{Adapter}(\cdot)$ denotes the Convpass adapter module \cite{jie_convolutional_2024}. By training only the adapter parameters while keeping the ViT backbone frozen, we reduce the number of trainable parameters and minimize the risk of overfitting. This makes the model more suitable for domain adaptation in medical imaging.

The goal of CAT is to learn domain-invariant feature representations so that samples from the same class, even when originating from different domains, are close together in the feature space, while samples from different classes are kept far apart. To achieve this, we train the adapters using SCL, which encourages intra-class compactness and inter-class separation. The loss function is defined as:
\begin{equation}
\label{eq:scl}
\mathcal{L}_{\text{scl}} = \sum_{i \in \mathcal{I}} \frac{-1}{|\mathcal{P}(i)|} \\ \sum_{p \in \mathcal{P}(i)} \log \frac{\exp(z_i \cdot z_p / \tau)}{\sum\limits_{a \in \mathcal{I} \setminus \{i\}} \exp(z_i \cdot z_a / \tau)},
\end{equation}
where $\mathcal{I}$ represents the index set of samples in a mini-batch, $z_i = f(x_i;\theta^{S})$ denotes the feature embedding of sample $i$ extracted by the backbone parameterized by $\theta^{S}$, and $\mathcal{P}(i)$ corresponds to the set of positive samples that share the same class as $i$. The denominator includes all samples except $i$ to normalize the contrastive similarity scores. The temperature parameter $\tau$ controls the sharpness of similarity scores to adjust the impact of different distances in the representation space. The similarity between embeddings is computed using a dot product to ensure that positive pairs are pulled closer together while negative pairs are pushed further apart.

Given that SCL forms positive pairs within each mini-batch, severe class and domain imbalance can bias positive pair construction toward majority groups, thereby reducing the contribution of minority classes or under-represented domains. To mitigate this issue during CAT training, we assign each training sample a weight based on its domain and class frequency, and use these weights to construct mini-batches through weighted random sampling.

Let the pooled source dataset used in CAT contain $N$ samples from $M$ source domains and $C$ classes. For each domain $i \in \{1,\dots,M\}$, let $N_i$ denote the number of samples in domain $i$, and let $n_{ij}$ denote the number of samples of class $j \in \{1,\dots,C\}$ within domain $i$. We compute the domain weight and the class weight (within each domain) as inverse-frequency terms:
\begin{equation}
\textit{w}_d^{i} = \frac{N}{N_i},
\end{equation}
\begin{equation}
\textit{w}_c^{ij} = \frac{N_i}{n_{ij}}.
\end{equation}
For a sample that belongs to domain $i$ and class $j$, the final sampling weight is defined as:
\begin{equation}
\label{eq:sample_weights}
\textit{w}^{ij} = \textit{w}_d^{i}\, \textit{w}_c^{ij}.
\end{equation}
Substituting the definitions above yields:
\begin{equation}
\textit{w}^{ij} = \frac{N}{N_i}\cdot\frac{N_i}{n_{ij}} = \frac{N}{n_{ij}},
\end{equation}
which shows that the effective weight is proportional to the inverse frequency of the corresponding $(\text{domain } i,\text{class } j)$ combination. These weights are then used to construct mini-batches via weighted random sampling, so that samples from rare domain--class pairs are selected more frequently during CAT training. The training process for CAT is shown in Algorithm~\ref{alg:cat}.

\begin{algorithm}[htbp]
\caption{Contrastive Adapter Training (CAT)}
\label{alg:cat}
\begin{algorithmic}[1]
\REQUIRE Source datasets $\{\mathcal{D}^S_i\}_{i=1}^{M}$, ViT backbone parameters $\theta_{\omega}^{S}$, temperature $\tau$
\ENSURE Frozen feature extractor $\theta^{S}=\{\theta_{\omega}^{S},\theta_{\omega'}^{S}\}$ with trained adapters

\STATE Initialize $\theta_{\omega'}^{S}$ (adapters); freeze $\theta_{\omega}^{S}$
\STATE Compute sampling weights $\textit{w}^{ij}$ from the pooled source set $\cup_{i=1}^{M}\mathcal{D}^S_i$ (Eq.~(\ref{eq:sample_weights}))
\FOR{each mini-batch $\mathcal{B}=\{(x_j,y_j,d_j)\}$ sampled from $\cup_{i=1}^{M}\mathcal{D}^S_i$ using $\textit{w}^{ij}$}
    \STATE Extract embeddings $z_j = f(x_j;\theta^{S})$ with frozen $\theta_{\omega}^{S}$ and trainable $\theta_{\omega'}^{S}$
    \FOR{each anchor $j \in \mathcal{B}$}
        \STATE Construct the positive set $\mathcal{P}(j)=\{k\in\mathcal{B}\setminus\{j\} \mid y_k=y_j\}$
    \ENDFOR
    \STATE Compute the supervised contrastive loss $\mathcal{L}_{\text{scl}}$ using Eq.~\ref{eq:scl} (temperature $\tau$)
    \STATE Update $\theta_{\omega'}^{S}$ by backpropagating $\nabla_{\theta_{\omega'}^{S}}\mathcal{L}_{\text{scl}}$
\ENDFOR
\STATE \textbf{return} $\theta^{S}=\{\theta_{\omega}^{S},\theta_{\omega'}^{S}\}$
\end{algorithmic}
\end{algorithm}

{\color{black}After training the backbone with CAT, the source models are trained independently.} To promote efficient learning and reduce overfitting, the trained backbone is kept frozen. For each source domain $i$, we attach a domain-specific bottleneck with parameters $\theta_{\psi}^{i}$ and a classification head with parameters $\theta_{\phi}^{i}$ on top of the frozen backbone, and optimize these layers using the binary cross-entropy loss, as follows:
\begin{equation} 
\mathcal{L}_{BCE} = - \frac{1}{N_i} \sum_{i=1}^{N_i} \left[ y_{ij} \log \hat{y}_{ij} + (1 - y_{ij}) \log (1 - \hat{y}_{ij}) \right],
\end{equation}
where $\hat{y}_{ij}$ denotes the predicted probability for the positive class. 
Once the source models are trained using their respective data, direct access to source data is no longer available due to privacy constraints. 

\subsection{Consensus Knowledge Distillation}
\label{sec:CKD}
CKD adapts a student model to the unlabeled target domain by treating the trained source models as a set of frozen teachers that provide soft supervisory signals. Rather than relying on a single teacher, CKD aggregates the predictions of all $M$ teachers to obtain a consensus target, which helps reduce the influence of any individual source-specific bias and provides a more stable supervision signal under domain shift. Moreover, using soft labels preserves inter-class similarity information and provides smoother gradients than hard pseudo-labels. This also mitigates hard-label bias when the consensus prediction is uncertain or incorrect (e.g., when teachers disagree), since the student is encouraged to match the full probability distribution rather than commit to a potentially erroneous one-hot target.

Given an unlabeled target sample $x^T_j$, each teacher $f_i$ produces a logit vector $a^i_j=f_i(x^T_j)$. We form a consensus logit representation by averaging teacher logits,
\begin{equation}
\label{eq:con_eq}
\bar{a}_j=\frac{1}{M}\sum_{i=1}^{M} a^i_j,
\end{equation}
which provides a single soft supervisory signal that reflects agreement across teachers for sample $j$. The student model $f_s(\cdot)$ is then optimized to align its predictive distribution with this consensus by minimizing the KL divergence
\begin{equation}
\mathcal{L}_{\text{CKD}}=\frac{1}{N_T}\sum_{j=1}^{N_T} D_{\text{KL}}\big(\sigma(\bar{a}_j)\ \parallel\ \sigma(f_s(x^T_j))\big),
\label{eq:l_ckd}
\end{equation}
where $\sigma(\cdot)$ denotes the softmax function, $N_T$ is the number of target samples, and $D_{\text{KL}}(\cdot\parallel\cdot)$ is defined as:
\begin{equation}
\label{eq:kl_divergence}
D_{\text{KL}}(P\parallel Q)=\sum_{c=1}^{C} P_c\log\frac{P_c}{Q_c},
\end{equation}
where $P_c=\sigma(\bar{a}_j)_c$ and $Q_c=\sigma(f_s(x^T_j))_c$ denote the class-$c$ probabilities. By training against the consensus soft target $\sigma(\bar{a}_j)$, the student distills shared knowledge across multiple sources while avoiding dependence on a single teacher, thereby improving robustness to heterogeneous source domains. Since adaptation relies only on teacher outputs for target samples and does not require access to source-domain images, CKD aligns with source-data-free deployment constraints in medical applications. The CKD procedure is summarized in Algorithm~\ref{alg:ckd}.

\begin{algorithm}[htbp]
\caption{Consensus Knowledge Distillation (CKD)}
\label{alg:ckd}
\begin{algorithmic}[1]
\REQUIRE Trained source models $\{f_1, f_2, ..., f_M\}$, target data $\mathcal{D}^T$
\ENSURE Adapted student model $f_s$
\STATE Randomly initialize student model $f_s$
\FOR{each mini-batch $\{x_j^T\}$ from $\mathcal{D}^T$}
    \FOR{each $x_j^T$ in the mini-batch}
        \FOR{$i = 1$ to $M$}
            \STATE Compute teacher logits: $a_j^i = f_i(x_j^T)$
        \ENDFOR
        \STATE Compute consensus logits $\bar{a}_j$ using Eq.~\ref{eq:con_eq}
    \ENDFOR
    \STATE Compute $\mathcal{L}_{\text{CKD}}$ using Eq.~\ref{eq:l_ckd} and update $f_s$ by backpropagation
\ENDFOR
\STATE \textbf{return} Trained student model $f_s$
\end{algorithmic}
\end{algorithm}

\subsection{Theoretical Justification}
\label{sec:theory_ckd}
To show why consensus distillation can provide a stable supervisory signal under domain shift, a brief analysis is provided under standard assumptions.

\begin{assumption}
For an input image $x$ and its corresponding class label $y$, let $\ell(x)\in\mathbb{R}^{C}$ denote a latent logit representation such that $\sigma(\ell(x))=P(y\mid x)$. Each teacher $f_i$ produces a logit vector that can be shown as:
\begin{equation}
a_j^i = \ell(x_j^T) + \epsilon_i(x_j^T), \quad \forall i\in\{1,\dots,M\},
\end{equation}
where $\epsilon_i(\cdot)$ captures the teacher-specific logit error on the target domain.
\end{assumption}

\begin{lemma}
If the teacher logit errors have bounded second moment and are not perfectly correlated, then the consensus logit $\bar{a}_j$ reduces the variance of the aggregated logit error.
\end{lemma}

\begin{proof}
Using Assumption~1 and the definition of the consensus logit in Eq.~\ref{eq:con_eq}, it follows that:
\begin{equation}
\bar{a}_j = \frac{1}{M}\sum_{i=1}^{M} a_j^i
= \ell(x_j^T) + \bar{\epsilon}(x_j^T),
\end{equation}
where
\begin{equation}
\bar{\epsilon}(x_j^T)=\frac{1}{M}\sum_{i=1}^{M}\epsilon_i(x_j^T).
\end{equation}

The mean-squared magnitude of the aggregated error satisfies:
\begin{equation}
\mathbb{E}\left[\|\bar{\epsilon}(x_j^T)\|_2^2\right]
=\frac{1}{M^2}\sum_{i=1}^{M}\sum_{k=1}^{M}\mathbb{E}\left[\langle \epsilon_i(x_j^T),\epsilon_k(x_j^T)\rangle\right].
\end{equation}
When the cross-correlations $\mathbb{E}[\langle \epsilon_i,\epsilon_k\rangle]$ for $i\neq k$ are limited (i.e., errors are not perfectly aligned), the above quantity decreases as $M$ increases, indicating that $\bar{a}_j$ is a more stable logit-level supervisory signal than $a_j^i$ from any single teacher.
\end{proof}

\begin{lemma}
For a fixed consensus logit $\bar{a}_j$, minimizing $D_{\text{KL}}(\sigma(\bar{a}_j)\parallel q_j)$ with respect to the student predictive distribution $q_j$ yields the unique optimum $q_j=\sigma(\bar{a}_j)$.
\end{lemma}

\begin{proof}
For any fixed $\bar{a}_j$, $D_{\text{KL}}(\sigma(\bar{a}_j)\parallel q_j)$ is minimized when $q_j$ matches the target distribution $\sigma(\bar{a}_j)$.
\end{proof}

\noindent Combining Lemma~1 and Lemma~2, training the student by minimizing
$D_{\text{KL}}(\sigma(\bar{a}_j)\parallel \sigma(f_s(x_j^T)))$ encourages the student to match a consensus target whose underlying logit error can be less variable than individual teacher logits when teachers exhibit complementary errors. This provides intuition for why CKD can improve robustness to heterogeneous source domains and domain shift.

\section{Experiments and Results}
\label{sec:experiments}
\subsection{Datasets} 
In this study, we utilized five publicly available datasets for skin lesion analysis: ISIC 2018 \cite{codella_skin_2018, tschandl_ham10000_2018}, ISIC 2020 \cite{rotemberg_patient-centric_2021}, PAD-UFES-20 \cite{pacheco_pad-ufes-20_2020}, Derm7pt-Derm \cite{kawahara_seven-point_2019}, and Derm7pt-Clinic \cite{kawahara_seven-point_2019}. Images of melanoma and nevus were selected from each dataset because these two classes are the most challenging to differentiate and are the most clinically significant for skin-lesion diagnosis. In addition, all of the datasets used in this study contain samples from these two classes, ensuring consistency across domains.

The Derm7pt dataset is partitioned into two subsets which are Derm7pt‑Derm (dermoscopic images) and Derm7pt‑Clinic (clinical images)\cite{wang_achieving_2024}. This split is necessary because the subsets come from different imaging modalities. Consequently, they exhibit distinct data distributions that influence the domain‑adaptation process. The number of melanoma and nevus samples in each dataset are summarized in Table \ref{tab:datasets}. As it can be seen, the datasets exhibit substantial class and sample-size imbalance, closely resembling deployment scenarios in clinical practice.

\begin{table}[htbp]
\centering
\renewcommand{\arraystretch}{1.2} 
\caption{Summary of the number of samples in skin lesion datasets used in the study.}
\label{tab:datasets}
\begin{tabular}{l|c|c}
\toprule
\textbf{Dataset}        & \textbf{Melanoma} & \textbf{Nevus} \\ \midrule
ISIC 2018 \cite{codella_skin_2018, tschandl_ham10000_2018}  & 1113 & 6705 \\
ISIC 2020 \cite{rotemberg_patient-centric_2021}             & 584  & 5193 \\
PAD-UFES-20 \cite{pacheco_pad-ufes-20_2020}                 & 52   & 244  \\
Derm7pt-Derm \cite{kawahara_seven-point_2019}               & 90   & 256  \\
Derm7pt-Clinic \cite{kawahara_seven-point_2019}             & 90   & 256  \\ \bottomrule
\end{tabular}
\end{table}

\begin{figure}[htbp]
  \centering
  \includegraphics[width=0.7\textwidth]{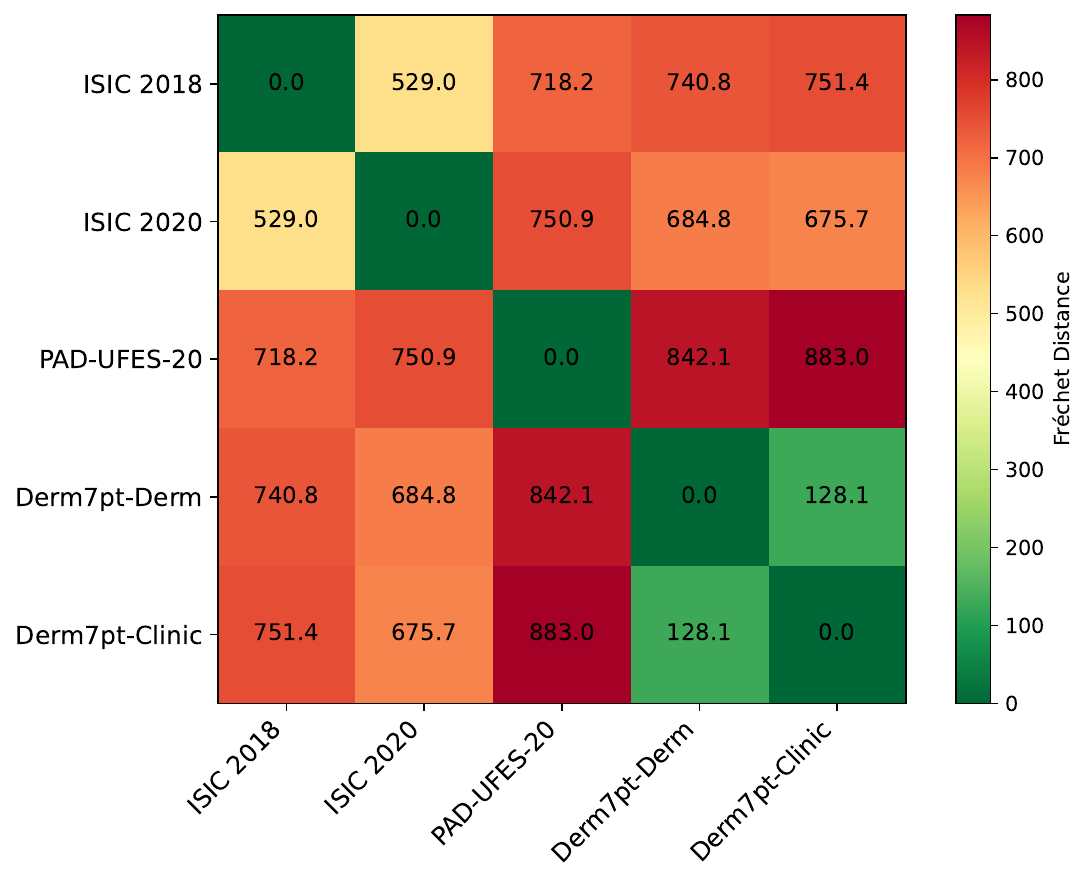}
  \caption{{\color{black}Pairwise Fréchet Distance between feature distributions of the five skin lesion datasets, where higher values indicate greater domain shift. The substantial distances observed across dataset pairs confirm the presence of significant inter-domain variability. This result motivates the use of domain adaptation.}}
  \label{fig:domain_gap}
\end{figure}

The selected datasets reflect real-world variability in imaging modalities, acquisition settings, and clinical sources, that introduces natural domain shifts essential for evaluating MSFDA methods. Their independent collection aligns with source-free constraints, where source data cannot be shared. To further quantify inter-dataset domain shift, Figure~\ref{fig:domain_gap} reports the pairwise Fréchet Distance between feature distributions. This highlights the significant visual and statistical domain shift between datasets used in this study.

\subsection{Implementation Details}
For all experiments, we use a batch size of 32 across all datasets to maintain consistency. ViT-B/16 pre-trained on ImageNet is used as the backbone across all experiments. The source models are trained for 20 epochs to ensure sufficient learning from the data, while the student model is trained for 15 epochs using the proposed CKD framework. The learning rate is set to $3 \times 10^{-2}$, with a warmup phase of 5\% of the total training steps to stabilize the optimization process during the initial epochs. SGD is the optimizer, configured with a momentum of 0.9 to enhance convergence by maintaining directional consistency during updates. $\tau$ in Eq. \ref{eq:scl} is set to 0.2 based on the details provided in Section \ref{sec:tau_effect}.
The bottleneck of our method contains four fully connected layers comprising 2048, 1024, 512, and 256 nodes, respectively. The head consists of a single fully connected layer. We treat each dataset as an independent domain and employ a leave-one-domain-out protocol, where, in each experimental setting, one domain is selected as the unlabeled target and all other domains are used as labeled sources.

\begin{figure}[htbp]
  \centering
    \includegraphics[width=0.9\textwidth]{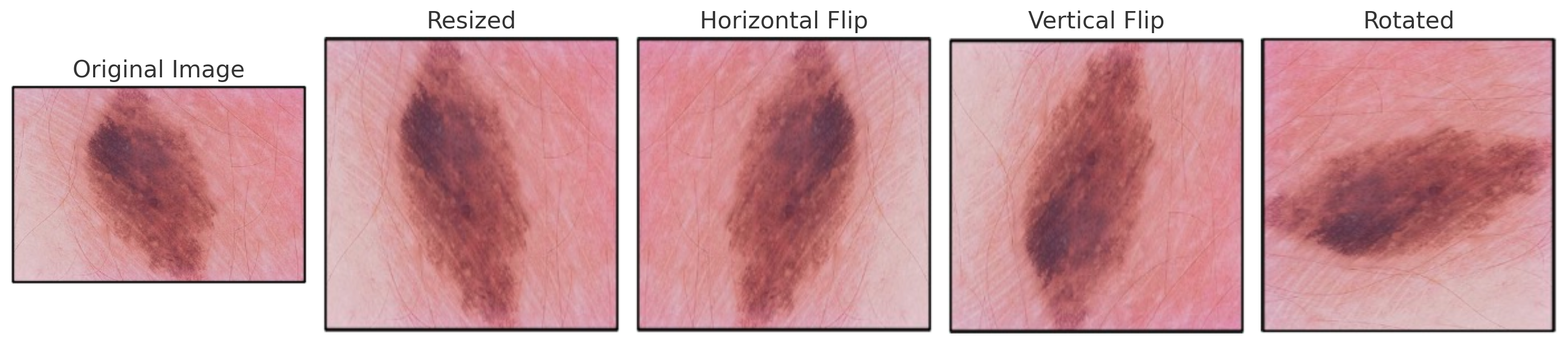}
  \caption{Demonstration of data augmentations applied to a skin lesion images in our proposed method.}
  \label{fig:augmentation}
\end{figure}

\textbf{Data Augmentation.}
To enhance the robustness and generalization capability of our model, we apply a series of data augmentation techniques during the training process (see Figure \ref{fig:augmentation}). These augmentations simulate diverse imaging conditions, which are common in real-world scenarios, thereby enabling the model to better handle variability in the data. Specifically, we use resizing to ensure uniform input dimensions, random horizontal and vertical flips to introduce variability in orientation, and random rotations within a range of 0 to 90 degrees to account for angular variations in imaging setups. These transformations are implemented as part of a pre-processing pipeline.

\subsection{Baselines}
To ensure a comprehensive evaluation, we select several well-established state-of-the-art models as baselines. Specifically, we consider both single-source and multi-source UDA methods, as well as multi-source-free UDA methods. 
First, we include DANN \cite{csurka_domain-adversarial_2017} and ADDA \cite{tzeng_adversarial_2017}, two widely recognized UDA models. Since these models are designed for single-source domain adaptation, we combine all source datasets into a single dataset and use a separate dataset as the target domain to align with their training paradigm. This setup allows for a fair comparison of their effectiveness in a multi-source setting.  
Next, we consider MDAN \cite{zhao_adversarial_2018}, a well-known multi-source domain adaptation method that explicitly handles multiple source domains. 
For comparison with our proposed MSFDA approach, we select DECISION \cite{ahmed_unsupervised_2021}, Bi-ATEN \cite{li_agile_2024}, and EAAF \cite{pei_evidential_2024}, all of which are specifically designed for scenarios where source data is unavailable during adaptation. Since our method operates under similar privacy-preserving constraints, these baselines provide an appropriate benchmark for evaluating the effectiveness of our proposed framework.  
All models are evaluated on the same datasets and data distributions to ensure a fair comparison.

\subsection{Results}

\textbf{Evaluation.}
We evaluate all models using the Area Under the Receiver Operating Characteristic Curve (AUROC), a threshold-independent metric suited for imbalanced datasets like skin cancer diagnosis. AUROC measures the model’s ability to rank positive samples higher than negative ones by computing the area under the ROC curve, defined as:

\begin{equation}
\text{AUROC} = \int_{0}^{1} \text{TPR}(t) \, d \text{FPR}(t),
\end{equation}
where True Positive Rate (TPR) and False Positive Rate (FPR) are:

\begin{equation}
\text{TPR} = \frac{\text{TP}}{\text{TP} + \text{FN}}, \quad
\text{FPR} = \frac{\text{FP}}{\text{FP} + \text{TN}}.
\end{equation}

Unlike accuracy, AUROC remains robust to class imbalance. This makes the AUROC ideal for medical imaging tasks.

\textbf{Performance.}
Table \ref{table:results} presents the AUROC scores for CAT-CKD and the competing baselines. {\color{black}To further illustrate the severity of domain shift across datasets, we include a Source Only baseline, in which all source domains are pooled and a single model is trained using standard cross-entropy without any domain adaptation.} The results demonstrate that CAT-CKD achieves strong performance across datasets in the MSFDA setting for skin cancer diagnosis. Importantly, the framework operates without access to source data during adaptation, thereby preserving data privacy. Despite this constraint, the method maintains robust generalization across domains, supporting its suitability for real-world clinical deployment. Notably, these results are obtained with a substantially reduced number of trainable parameters.

\begin{table*}[htbp]
\caption{AUROC comparison of domain adaptation models on skin lesion datasets. Best results are shown in bold.}
\label{table:results}
\centering
\resizebox{1\textwidth}{!}{
\begin{tabular}{l|cc|ccccc|c}
\toprule
\textbf{Model} & \textbf{MS} & \textbf{SF} & \textbf{ISIC 2018} & \textbf{ISIC 2020} & \textbf{PAD-UFES-20} & \textbf{Derm7pt-Derm} & \textbf{Derm7pt-Clinic} & \textbf{Avg.} \\
\midrule
Source Only & $\times$ & $\times$ & $78.3 \pm 4.8$ & $54.1 \pm 6.5$ & $45.8 \pm 7.4$ & $50.8 \pm 5.1$ & $48.6 \pm 2.3$ & $55.5 \pm 5.2$ \\
ADDA \cite{tzeng_adversarial_2017} & $\times$ & $\times$ & $92.1 \pm 0.8$ & $83.0 \pm 0.4$ & $72.5 \pm 0.7$ & $77.2 \pm 0.3$ & $72.2 \pm 0.6$ & $79.4 \pm 0.5$ \\
DANN \cite{csurka_domain-adversarial_2017} & $\times$ & $\times$ & $88.7 \pm 0.6$ & $81.7 \pm 0.6$ & $79.9 \pm 0.7$ & $80.2 \pm 0.4$ & $78.3 \pm 0.5$ & $81.8 \pm 0.6$ \\ 
MDAN \cite{zhao_adversarial_2018} & $\checkmark$ & $\times$ & $84.9 \pm 0.7$ & $75.1 \pm 0.6$ & $79.8 \pm 0.9$ & $83.7 \pm 0.5$ & $77.1 \pm 0.7$ & $80.1 \pm 0.7$ \\
DECISION \cite{ahmed_unsupervised_2021} & $\checkmark$ & $\checkmark$ & $76.3 \pm 1.5$ & $58.7 \pm 0.7$ & $77.8 \pm 1.1$ & $79.1 \pm 0.6$ & $56.2 \pm 0.8$ & $69.6 \pm 0.9$ \\
EAAF \cite{pei_evidential_2024} & $\checkmark$ & $\checkmark$ & $80.8 \pm 0.7$ & $72.5 \pm 0.6$ & $78.5 \pm 0.5$ & $82.4 \pm 0.7$ & $79.7 \pm 0.4$ & $78.8 \pm 0.6$ \\
Bi-ATEN \cite{li_agile_2024} & $\checkmark$ & $\checkmark$ & $81.2 \pm 1.2$ & $86.5 \pm 0.8$ & $83.5 \pm 0.7$ & $86.5 \pm 0.5$ & $85.7 \pm 1.0$ & $84.3 \pm 0.4$ \\
CAT-CKD (Ours) & $\checkmark$ & $\checkmark$ & $90.8 \pm 0.2$ & $90.5 \pm 0.5$ & $75.8 \pm 0.6$ & $89.7 \pm 0.1$ & $84.0 \pm 0.5$ & $\textbf{86.1} \pm 0.4$ \\
\bottomrule
\end{tabular}
}
\end{table*}

\begin{figure*}[htbp]
\centerline{\includegraphics[width=1\textwidth]{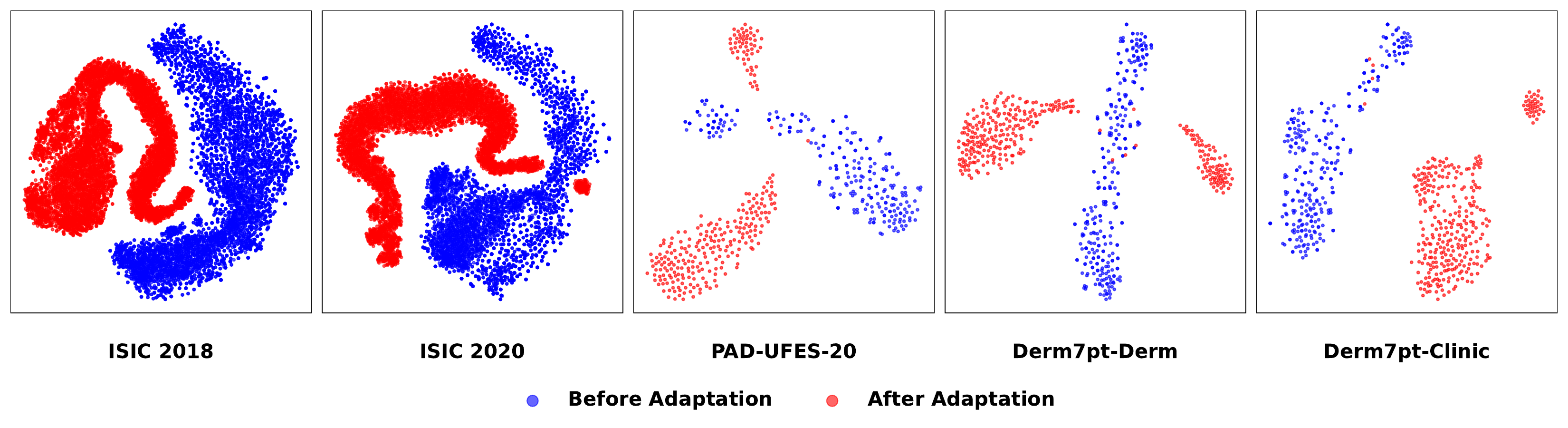}}
\caption{t-SNE visualization of target-domain features before and after adaptation with CAT-CKD.}
\label{fig:tsne}
\end{figure*}

To further illustrate the effectiveness of CAT-CKD, Figure~\ref{fig:tsne} presents t-SNE visualizations of the target-domain feature embeddings before and after adaptation. Following adaptation, the features exhibit markedly improved inter-class separability and intra-class compactness. The resulting clusters are more coherent and better aligned with class boundaries, suggesting that CAT-CKD successfully learns a discriminative and domain-consistent feature representation that enhances generalization on the target domain.

\textbf{Efficiency Analysis.}
Table~\ref{table:params} presents a comparison of trainable parameters between our proposed model and baselines. Existing methods, including EAAF~\cite{pei_evidential_2024} and DECISION~\cite{ahmed_unsupervised_2021}, involve training over 96M parameters, while others like DANN~\cite{csurka_domain-adversarial_2017} and ADDA~\cite{tzeng_adversarial_2017} still require tens of millions of parameters to be updated.
In contrast, our parameter-efficient design significantly reduces the number of trainable parameters by integrating lightweight adapters into a frozen ViT backbone. The full ViT model contains over 90 million parameters, but during the backbone training phase using CAT, only the adapter modules are updated—resulting in just 0.3 million trainable parameters. In the subsequent source model training and target adaptation stages, only the bottleneck and classification head are trained, with the total number of trainable parameters after CAT approximately 4.3 million. This represents a substantial reduction compared to fully fine-tuned baselines. The reduction in trainable parameters not only improves computational and memory efficiency, but also helps mitigate overfitting.

\begin{table}[htbp]
\centering
\renewcommand{\arraystretch}{1.2} 
\caption{Comparison of the trainable parameters between our proposed model and baseline models is presented.}
\label{table:params}
\resizebox{0.5\textwidth}{!}{
\begin{tabular}{lc}
\toprule
\textbf{Model Configuration} & \textbf{\# Parameters} \\ \midrule
EAAF \cite{pei_evidential_2024} & 98.2M \\
DECISION \cite{ahmed_unsupervised_2021} & 96.1M \\ 
DANN \cite{csurka_domain-adversarial_2017} & 53.0M \\ 
ADDA \cite{tzeng_adversarial_2017} & 26.0M \\ 
MDAN \cite{zhao_adversarial_2018} & 11.1M \\ 
Bi-ATEN \cite{li_agile_2024} & 10.6M \\
CAT-CKD (Full model) & 4.3M \\ \bottomrule
\end{tabular}
}
\end{table}

\textbf{Discussion on Results.}
The experimental results demonstrate the practical effectiveness of the proposed CAT-CKD framework in the context of MSFDA for skin cancer diagnosis. By adopting a parameter-efficient tuning strategy, the ViT backbone is adapted using lightweight adapters, enabling efficient optimization. This results in a robust and domain-invariant feature extractor capable of strong cross-domain generalization. Notably, the framework maintains competitive performance across diverse datasets without requiring access to source data during target adaptation, which is particularly important for privacy-sensitive medical applications. In addition, the significant reduction in trainable parameters improves computational efficiency and enhances suitability for deployment in resource-constrained clinical environments.

{\color{black}
PAD-UFES-20 presents a uniquely challenging adaptation scenario, as confirmed by the pairwise Fréchet Distance matrix (Figure~\ref{fig:domain_gap}), where it exhibits the largest domain distances from all other datasets. The consistent performance drop across methods indicates that the difficulty arises from the severity of the domain gap rather than from CAT-CKD. Despite this, CAT-CKD achieves the highest average AUROC of 86.1\% and outperforms all competing methods.

Nevertheless, adaptation to target domains with large domain gaps remains challenging, especially when imaging modalities or acquisition conditions differ substantially. Moreover, while the framework extends to multi-class or multi-label settings with minimal modifications, constructing cross-domain datasets with consistent label definitions and realistic domain shifts remains a practical challenge.}

\subsection{Ablation Study}
In this section, we conduct an ablation study to evaluate the effectiveness of each component of the CAT-CKD framework. As such, we remove these components individually to assess their contributions. We first remove CAT and use the backbone in a frozen manner without the CAT-based finetuning step. We then evaluated fine-tuning the entire backbone instead of applying the CAT-based parameter-efficient adaptation. This full fine-tuning strategy resulted in inferior performance, due to overfitting on limited data and catastrophic forgetting of pretrained representations. To eliminate CKD, we supervise the student model using logits from a randomly selected single source model instead of leveraging the ensemble. For removing KL divergence, we substitute it with a standard cross-entropy loss. As shown in Table~\ref{table:abl_std}, removing each component lead to notable performance degradation. These results confirm that all components in our proposed method are essential to the framework’s success, as removing either results in a substantial drop in AUROC, thereby validating the proposed design.

\begin{table*}[htbp]
\caption{Ablation study evaluating the impact of key components in our framework. We report AUROC scores across five target datasets and the average. Best results are shown in bold.}
\label{table:abl_std}
\centering
\resizebox{1\textwidth}{!}{
\begin{tabular}{l|ccccc|c}
\toprule
\textbf{Model} & \textbf{ISIC 2018} & \textbf{ISIC 2020} & \textbf{PAD-UFES-20} & \textbf{Derm7pt-Derm} & \textbf{Derm7pt-Clinic} & \textbf{Avg.} \\
\midrule
Ours & 90.8 & 90.5 & 75.8 & 89.7 & 84.0 & \textbf{86.1} \\
Ours w/o CAT & 87.5 & 90.1 & 72.3 & 78.6 & 78.0 & 81.3 \\
Ours w/o CAT, w/ Full backbone finetuning & 78.6 & 90.1 & 64.5 & 65.9 & 75.4 & 74.9 \\
Ours w/o CKD & 86.0 & 59.5 & 53.4 & 73.2 & 74.2 & 69.3 \\
Ours w/o KL Divergence & 85.7 & 61.2 & 53.3 & 72.1 & 73.5 & 69.2 \\
\bottomrule
\end{tabular}
}
\end{table*}

\subsection{Sensitivity Analysis}
In this section, we perform a sensitivity analysis to evaluate how hyperparameter settings and architectural components influence our model’s performance.

\subsubsection{Effect of CAT on Attention Maps}
\begin{figure*}[htbp]
\centerline{\includegraphics[width=1\textwidth]{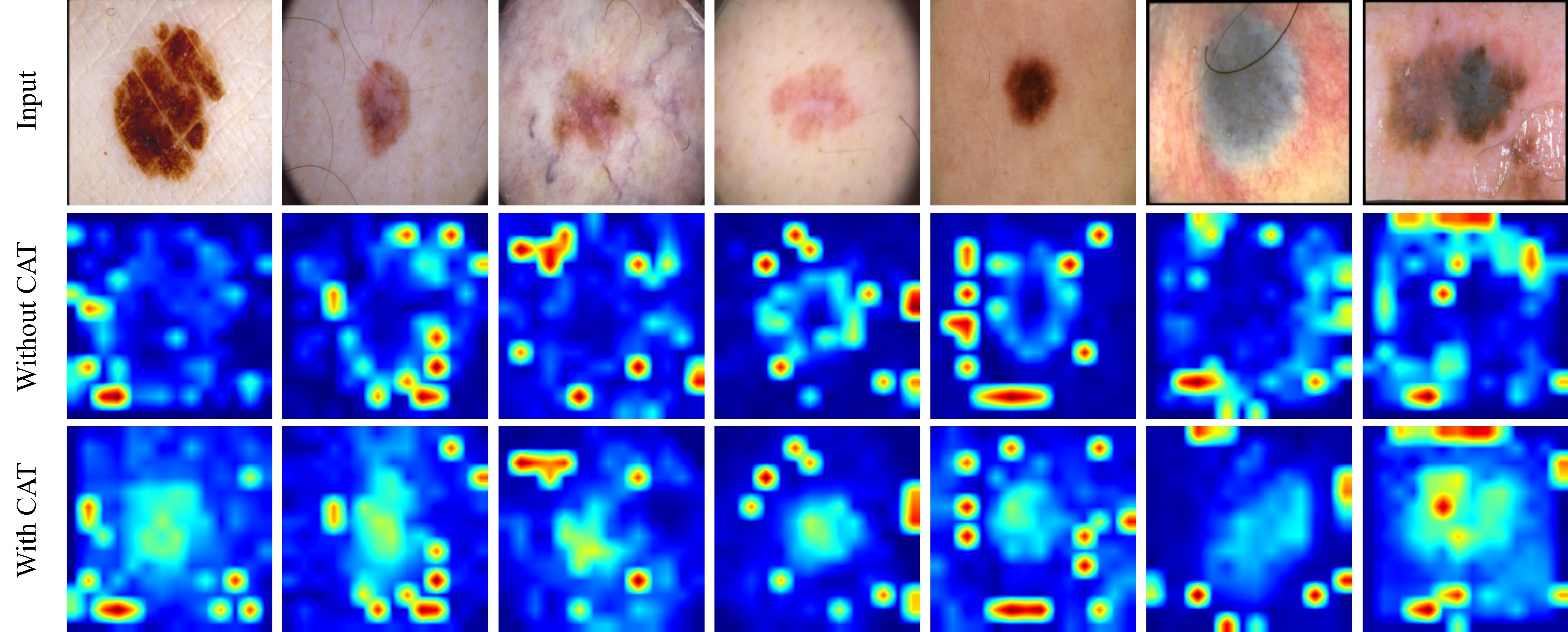}}
\caption{Comparison of attention maps generated by the ViT backbone with and without CAT.}
\label{fig:attention_maps}
\end{figure*}

{\color{black}
Figure~\ref{fig:attention_maps} shows attention maps generated by the backbone, with and without CAT. Without CAT, attention is broadly dispersed across irrelevant background regions with little to no focus on the lesion area, indicating poor feature localization. Introducing CAT noticeably concentrates attention toward lesion regions, demonstrating that contrastive adapter training encourages the model to attend to more semantically relevant areas. While perfect lesion localization is not expected from ViT attention maps due to their global patch-relationship nature, the key observation is that CAT shifts the attention distribution toward more focused and semantically meaningful patterns, which indicates improved domain-invariant feature learning.}

\subsubsection{Soft-label vs. Hard-label Knowledge Distillation}
To evaluate the impact of supervision type during distillation, we compare the proposed soft-label CKD with a hard-label variant in which the consensus prediction is converted into one-hot pseudo-labels and the student is trained using cross-entropy loss. As shown in Table~\ref{table:soft_vs_hard}, soft-label CKD achieves a higher average AUROC and delivers more robust performance on challenging targets such as PAD-UFES-20. This behavior arises because hard-label distillation propagates teacher errors directly when the consensus prediction is incorrect and enforces an overconfident, potentially noisy target. In contrast, soft-label distillation trains the student using the full predictive distribution, which preserves uncertainty and inter-class similarity information from multiple teachers.

\begin{table}[htbp]
\caption{Comparison between soft-label and hard-label knowledge distillation.}
\label{table:soft_vs_hard}
\centering
\resizebox{1\textwidth}{!}{
\begin{tabular}{l|ccccc|c}
\toprule
\textbf{Model} & \textbf{ISIC 2018} & \textbf{ISIC 2020} & \textbf{PAD-UFES-20} & \textbf{Derm7pt-Derm} & \textbf{Derm7pt-Clinic} & \textbf{Avg.} \\
\midrule
Hard-label CKD        & 87.4 & 86.9 & 71.3 & 87.7 & 83.7 & 83.4 \\
Soft-label CKD (Ours) & 90.8 & 90.5 & 75.8 & 89.7 & 84.0 & \textbf{86.1} \\
\bottomrule
\end{tabular}
}
\end{table}

\subsubsection{Convpass vs. Other Adapter Models}
To assess the effect of adapter architecture on the proposed method, we compare the ConvPass adapter with two widely used alternatives. The Houlsby adapter~\cite{houlsby_parameter-efficient_2019} inserts bottleneck modules after both the attention and feed-forward sublayers within each transformer block. In contrast, the Pfeiffer adapter~\cite{pfeiffer_adapterfusion_2021} adopts a more lightweight design by placing a single adapter after the feed-forward layer only. Both variants employ a linear down-projection, non-linearity, and up-projection structure, but differ in their placement and overall expressiveness (see Table~\ref{table:adapters}).

\begin{table}[htbp]
\caption{Comparison between different adapter models.}
\label{table:adapters}
\centering
\resizebox{1\textwidth}{!}{
\begin{tabular}{l|ccccc|c}
\toprule
\textbf{Model} & \textbf{ISIC 2018} & \textbf{ISIC 2020} & \textbf{PAD-UFES-20} & \textbf{Derm7pt-Derm} & \textbf{Derm7pt-Clinic} & \textbf{Avg.} \\
\midrule
CAT(w/ Houlsby~\cite{houlsby_parameter-efficient_2019})-CKD  & 90.7 & 73.7 & 68.0 & 91.2 & 84.6 & 81.6 \\
CAT(w/ Pfeiffer~\cite{pfeiffer_adapterfusion_2021})-CKD & 89.7 & 77.9 & 74.1 & 91.7 & 81.0 & 82.8 \\
CAT(w/ Convpass~\cite{jie_convolutional_2024})-CKD (Ours) & 90.8 & 90.5 & 75.8 & 89.7 & 84.0 & \textbf{86.1} \\
\bottomrule
\end{tabular}
}
\end{table}

\subsubsection{Effect of Different Consensus Label Generation Strategies}
We evaluate different strategies for generating consensus pseudo-labels from multiple teacher models during distillation. Under domain shift, variations in teacher confidence and reliability make the aggregation strategy a critical component of the distillation process.
As shown in Table~\ref{table:consensus}, in addition to uniform averaging (used in our method), we consider three confidence-aware alternatives: confidence-weighted averaging \cite{chen_semi-supervised_2018} (based on maximum softmax score), margin-based weighting \cite{sosea_marginmatch_2023} (using the top-1 vs. top-2 probability gap), and entropy-weighted averaging \cite{long_conditional_2018} (favoring low-entropy predictions). While these strategies aim to prioritize confident teachers, they rely on potentially miscalibrated confidence estimates. This can amplify errors from overconfident but incorrect teachers. Uniform averaging, by contrast, avoids such bias and provides a stable, balanced supervision signal across domains.

\begin{table}[htbp]
\caption{Comparison between different consensus label generation strategies.}
\label{table:consensus}
\centering
\resizebox{1\textwidth}{!}{
\begin{tabular}{l|ccccc|c}
\toprule
\textbf{Model} & \textbf{ISIC 2018} & \textbf{ISIC 2020} & \textbf{PAD-UFES-20} & \textbf{Derm7pt-Derm} & \textbf{Derm7pt-Clinic} & \textbf{Avg.} \\
\midrule
Confidence-weighted averaging \cite{chen_semi-supervised_2018} & 91.2 & 75.7 & 70.8 & 90.7 & 86.2 & 82.9 \\
Margin-based weighting \cite{sosea_marginmatch_2023} & 90.8 & 75.0 & 73.1 & 90.7 & 85.3 & 83.0 \\
Entropy-weighted averaging \cite{long_conditional_2018} & 89.1 & 84.4 & 72.2 & 89.6 & 84.1 & 83.9 \\
CKD (Ours) & 90.8 & 90.5 & 75.8 & 89.7 & 84.0 & \textbf{86.1} \\
\bottomrule
\end{tabular}
}
\end{table}

\subsubsection{Effect of \texorpdfstring{$\tau$}{tau} in Supervised Contrastive Learning}
\label{sec:tau_effect}
\begin{figure}[htbp]
  \centering
    \includegraphics[width=0.7\textwidth]{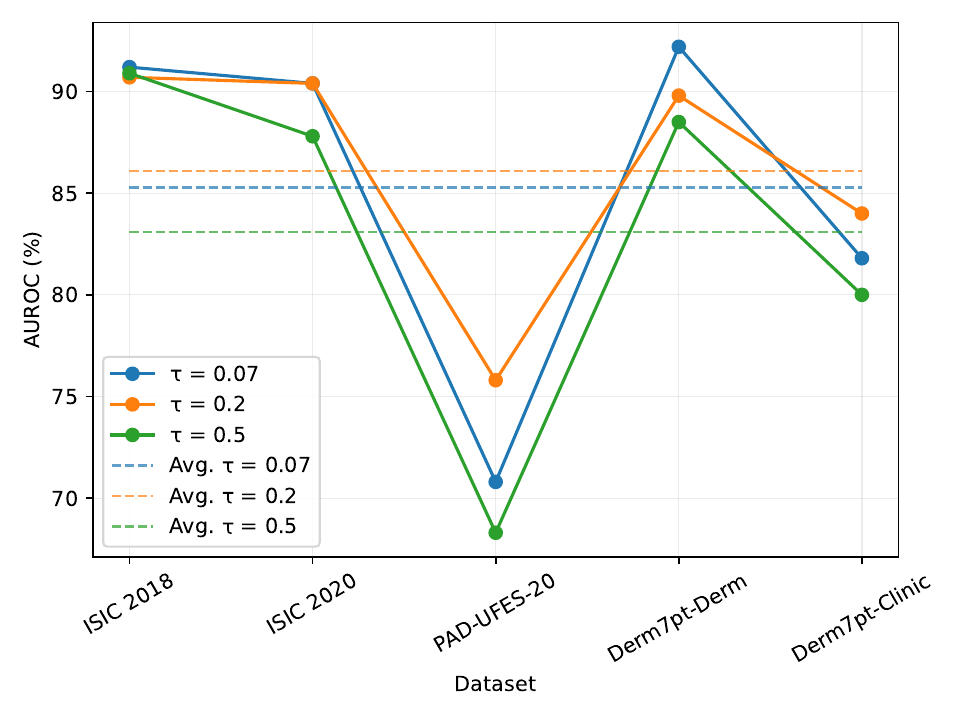}
  \caption{AUROC performance comparison for different temperature values.}
  \label{fig:tau}
\end{figure}

The temperature parameter $\tau$ in SCL modulates the concentration of the similarity distribution and thus determines the strength of separation between positive and negative pairs. Lower values yield sharper distributions that emphasize hard negatives, whereas higher values produce smoother distributions with reduced contrastive force. We tested three common values: 0.07, 0.2, and 0.5. As shown in Figure \ref{fig:tau}, $\tau = 0.2$ delivers the best performance by generating the most discriminative representations. In contrast, $\tau =$ 0.07 and 0.5  both diminish class separation, underscoring the importance of tuning $\tau$ for effective domain-invariant feature learning.

\section{Conclusion}
\label{sec:conclusion}
In this work, we proposed CAT-CKD, a novel MSFDA framework for skin cancer diagnosis that preserves patient privacy by eliminating the need for source-domain images during adaptation. First, CAT instills domain-invariant, class-discriminative representations into a ViT backbone through SCL and lightweight adapters to reduce overfitting while retaining pretrained knowledge. Next, CKD transfers ensemble knowledge from frozen source models to a student network on unlabeled target data via a KL-divergence objective. Extensive experiments on five public dermoscopic datasets demonstrate that our framework consistently outperforms both traditional UDA and state-of-the-art multi-source-free methods. Moreover, the adapter-based design slashes the number of trainable parameters and cuts computational overhead, enabling practical deployment in resource-constrained clinical settings. 

Future work will focus on developing a more efficient source-domain training phase to reduce complexity. We will also investigate the robustness of the framework under different attack settings and privacy constraints. In addition, extending the approach to federated multi-source domain adaptation will be explored to further enhance its applicability in distributed and privacy-sensitive environments.

\section*{Acknowledgment}
This research is partially funded by the NSERC CREATE TrustCAV program and the NSERC Discovery Grant program, and in part by the NSERC Canada Research Chair (CRC) Program.

\bibliographystyle{elsarticle-num} 
\bibliography{references}
\end{document}